%% file: main.tex
\documentclass[USenglish,oneside,twocolumn]{article}

\usepackage[utf8]{inputenc}
\usepackage[big]{dgruyter_NEW}
\usepackage{multirow}
\usepackage{adjustbox}

\graphicspath{ {./figures/} } 
\newcommand{\mx}[0]{\textsc{Mutant-X}}

\newcommand{\nf}[1]{\normalfont{#1}}

\begin{document}

  \author*[1]{Muhammad Haroon}

  \author[2]{Fareed Zaffar}

  \author[3]{Padmini Srinivasan}

  \author[4]{Zubair Shafiq}

  \affil[1]{University of California, Davis, E-mail: mharoon@ucdavis.edu}

  \affil[2]{Lahore University of Management Sciences, E-mail: fareed.zaffar@lums.edu.pk}

  \affil[3]{The University of Iowa, E-mail: padmini-srinivasan@uiowa.edu}

  \affil[4]{University of California, Davis, E-mail: zshafiq@ucdavis.edu}
  
  \title{\huge Avengers Ensemble! Improving Transferability of Authorship Obfuscation}

  \runningtitle{Avengers Ensemble! Improving Transferability of Authorship Obfuscation}

  \input{sections/abstract}

  \maketitle

  \input{sections/introduction}
  \input{sections/methodology}
  \input{sections/experimental-setup}

  \input{sections/results}

  \input{sections/related}
  \input{sections/conclusion}

  \bibliographystyle{plainnat}
  \bibliography{obfuscation}

\end{document}

%% file: sections/abstract.tex
\begin{abstract}
{
Stylometric approaches have been shown to be quite effective for real-world authorship attribution.
To mitigate the privacy threat posed by authorship attribution, researchers have proposed automated authorship obfuscation approaches that aim to conceal the stylometric artefacts that give away the identity of an anonymous document's author. 
Recent work has focused on authorship obfuscation approaches that rely on black-box access to an attribution classifier to evade attribution while preserving semantics. 
However, to be useful under a realistic threat model, it is important that these obfuscation approaches work well even when the adversary's attribution classifier is different from the one used internally by the obfuscator.  
Unfortunately, existing authorship obfuscation approaches do not transfer well to unseen attribution classifiers.
In this paper, we propose an ensemble-based approach for transferable authorship obfuscation. 
Our experiments show that if an obfuscator can evade an ensemble attribution classifier, which is based on multiple base attribution classifiers, it is more likely to transfer to different attribution classifiers.  
Our analysis shows that ensemble-based authorship obfuscation achieves better transferability because it combines the knowledge from each of the base attribution classifiers by essentially averaging their decision boundaries.
}
\end{abstract}

%% file: sections/introduction.tex
\section{Introduction}
Authorship obfuscation is the process of concealing stylometric pointers in a text document that may reveal the identity of its author. 
The problem has become increasingly relevant today considering the erosion of privacy due to recent advances in the performance of state-of-the-art authorship attribution approaches.
Sophisticated machine learning models can determine the author of a given text document \cite{juola2010empirical,stolerman2014breaking} using hand-crafted stylometric features \cite{abbasi2008writeprints, brennan2012adversarial, mcdonald2012use, clark2007algorithm, afroz2014doppelganger} or automated features such as word embeddings \cite{ruder2016character, howard2018universal}.
State-of-the-art authorship attribution approaches have achieved impressive results in a multitude of settings ranging from social media posts \cite{almishari2014stylometric, overdorf2016blogs, rajapaksha2017identifying} to large-scale settings with up to a 100,000 possible authors \cite{narayanan2012feasibility}.

The desire to maintain anonymity in this increasingly hostile environment motivates the need for effective authorship obfuscation methods.
Obfuscation approaches can be broadly divided into two groups: those that do not rely on feedback from an authorship attribution classifier and those that do require such feedback.

In the first group, there are a number of efforts especially from the PAN digital text forensics initiative \cite{panobfuscation}. 
These authorship obfuscation approaches mostly use rule based transformations (e.g., splitting or joining sentences) guided by some general criteria, such as moving the text towards some average point or moving it away from the author's writing patterns, text simplification, machine translation, etc. \cite{Keswani2016AuthorMT,castro2017AuthorMB, karadzhovPAN2016, Potthast2016AuthorOA}. 
These approaches generally struggle with achieving the appropriate trade off between evasion effectiveness and preserving text semantics.

In the second group, obfuscators that rely on access to an authorship attribution classifier are more relevant to our research.
In a seminal work, MacDonald et al. \cite{mcdonald2012use} proposed Anonymouth -- an obfuscator that relies on access to the attributor JStylo to guide manual text obfuscation.
A\textsuperscript{4}NT \cite{a4nt2016shetty} proposed a \textit{generative adversarial network} (GAN) based automated approach to obfuscation and also requires access to the attribution classifier.
More recently, \mx{} \cite{DBLP:journals/popets/MahmoodASSZ19} used a genetic algorithm and ParChoice \cite{parchoice2020} used combinatorial paraphrasing for automated obfuscation, and both require access to an attribution classifier.
These methods have shown promise in effectively evading attribution classifiers while reasonably preserving text semantics.

While prior authorship obfuscation methods can suitably trade off between evading attribution and preserving semantics, they do not work well when the adversary uses a different attribution classifier than the one used internally by the obfuscator \cite{DBLP:journals/popets/MahmoodASSZ19,parchoice2020,a4nt2016shetty}.
However, it is important that authorship obfuscators can protect the author's identity even when the adversary uses a different attribution classifier. 
In other words, obfuscation should transfer to previously unseen attribution classifiers.

The lack of transferability is essentially because of the mismatch between the obfuscator's internal classifier and the adversary’s attribution classifier.
To address the transferability issue, our key insight is that if an obfuscator can evade a meta-classifier, which is based on multiple base classifiers that target different feature subspaces, it is more likely to evade an unseen attribution classifier.
Building on this insight, we propose an ensemble-based approach for transferable authorship obfuscation. 
We explore the design space of the ensemble using different feature subspaces, base classifiers, and aggregation techniques. 
The experimental evaluation shows our ensemble based authorship obfuscation approach yields state-of-the-art transferability results.
We find that obfuscation by our ensemble approach achieves an 1.7$\times$ and 2.1$\times$ better transferability in terms of the attack success rate (ASR) than the baseline RFC and SVM attributors.
The ensemble achieves an average METEOR score of 0.36, which is comparable with the RFC at 0.42 and the SVM at 0.40.

We summarize our key contributions and findings as follows:

\begin{enumerate}

\item  We explore the problem of transferability of authorship obfuscation against unseen attribution classifiers. 

\item We propose an ensemble approach that consists of multiple base classifiers, each capturing different feature subspaces, to guide automated text obfuscation. 

\item We evaluate the evasion effectiveness, semantic preservation, and transferability of the ensemble obfuscator and show that it achieves much better transferability against unseen attribution classifiers than prior approaches.

\end{enumerate}

%% file: sections/methodology.tex
\section{Preliminaries \& Methods}

\begin{figure*}[t]
    \includegraphics[width=\textwidth,keepaspectratio]{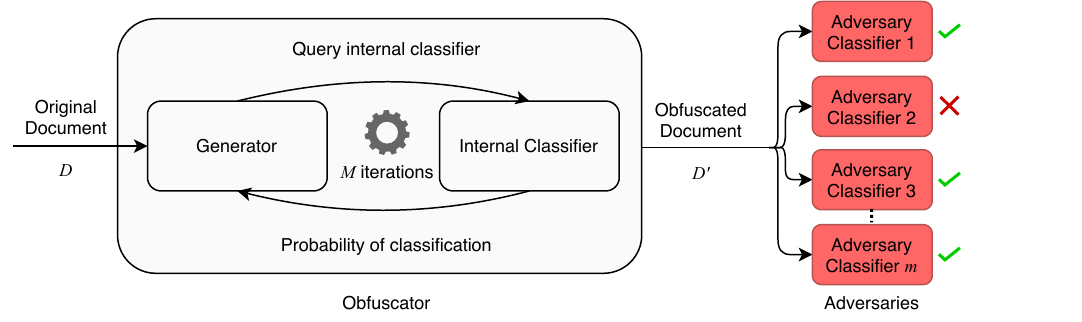}
    \caption{Overview of the threat model involving an obfuscator and multiple adversaries. The obfuscator relies on access to an internal attribution classifier. The obfuscator consists of two components: a generator and an obfuscator. The generator generates an obfuscation and queries the internal classifier for feedback on its probability of detection. They repeat this for $M$ iterations and then, if evaded, the final document is verified against the adversaries.}
    \label{fig:black-box}
\end{figure*}

\subsection{Authorship Attribution vs. Obfuscation}
Stylometry is the analysis of an author's writing style that helps distinguish them from other authors.
For example, \textit{writeprints} \cite{abbasi2008writeprints} is a well-known stylometric feature set that has been used to analyze the writing style for the sake for authorship attribution.
The primary goal of authorship obfuscation is to evade attribution by concealing such stylometric features in the document while retaining its original meaning.
Early approaches such as Anonymouth \cite{mcdonald2012use} highlighted the distinctive stylometric properties of the text that could then be modified by the user to evade attribution.
Follow up work at PAN-CLEF aimed to automatically obfuscate documents using simple predefined rules.
For example, Mansoorizadeh et al. \cite{mansoorizadehPAN2016} used WordNet to identify synonyms for the most commonly used words by the author and replaced them with  similar words.
Castro et al. \cite{castro2017AuthorMB} used sentence simplification techniques, such as replacing contractions with their expansions, to obfuscate a document.
Keswani et al. \cite{Keswani2016AuthorMT} used round-trip translation ($English \rightarrow German \rightarrow French \rightarrow English$) to obfuscate a document. 
While these automated approaches managed to evade attribution, they severely compromised the obfuscated text's semantics.
These approaches, rather unsuccessfully, navigate the trade-off between evading attribution and preserving semantics \cite{DBLP:journals/popets/MahmoodASSZ19}.

Recent work such as A\textsuperscript{4}NT \cite{a4nt2016shetty}, \mx \cite{DBLP:journals/popets/MahmoodASSZ19}, and ParChoice \cite{parchoice2020} employ more sophisticated adversarial obfuscation to evade authorship attribution classifiers. 
Their threat model assumes that the obfuscator can query the adversary's attribution classifier to guide obfuscation.
For example, A\textsuperscript{4}NT uses a \textit{generative adversarial network} (GAN) for obfuscation that requires white-box access to the adversary's attribution classifier.
\mx{} uses a genetic algorithm for obfuscation that requires black-box access to the adversary's attribution classifier.
ParChoice uses combinatorial paraphrasing for obfuscation that requires black-box access to adversary's attribution classifier.
While these obfuscation approaches achieve a better tradeoff between attribution evasion and preserving semantics, they all assume white/black box access to the the adversary's attribution classifier.
This key assumption limits their effectiveness in the real world because the adversary's attribution classifier might be different or unknown. 
For example, the evasion effectiveness of \mx{} drops drastically when the adversary uses a different attribution classifier than assumed by \mx{} \cite{DBLP:journals/popets/MahmoodASSZ19}.
Similarly, an adversarially retrained attribution classifier is resistant to obfuscation by ParChoice using the original classifier \cite{parchoice2020}.
This lack of \textit{transferability} to unseen attribution classifiers has major ramifications in the real world as the obfuscator's effectiveness is questionable when the adversary happens to use a different attribution classifier.
Figure \ref{fig:black-box} provides an overview of this threat model involving the obfuscator and multiple unseen adversaries.

\subsection{Problem Statement}
The obfuscator seeks to obfuscate the stylometric properties of an input document $D$ of author $A$ by modifying its text to produce an obfuscated document $D'$ such that the  attributor incorrectly classifies the obfuscated document to another author $A'\neq A$.
The state-of-the-art authorship obfuscators mainly consist of two components: a generator and an internal authorship attribution classifier $C$.
The generator modifies the input document based on some rules and queries the internal classifier to predict whether these modifications would degrade the likelihood of successful authorship attribution.
The two components work in tandem for $M$ iterations to progressively obfuscate the input document by generating new obfuscation samples and measuring the degradation in authorship attribution by $C$.
It is noteworthy that the adversary might use a different authorship attribution classifier $C'\neq C$.
There could, in fact, be multiple adversaries in this setting, with each using a different attribution classifier $C'$ than the obfuscator's internal classifier $C$.
The primary goal of the obfuscator is to obfuscate an input document $D \rightarrow D'$ using its internal classifier $C$ such that it evades attribution by the adversary classifier $C'$.
This problem is also referred to as \textit{transferability} in the field of adversarial machine learning. 

\subsection{Approach}

\medskip
\noindent \textbf{Intuition.}
The obfuscator relies on feedback from its internal classifier as a proxy to identify suitable transformations that can help evade attribution by the adversary's classifier.
These transformations essentially aim to move the document to the wrong side of the decision boundary, which partition different author classes, of the obfuscator's internal classifier.
%
%
Since these transformations are specific to the decision boundary of the obfuscator's internal classifier, they may not achieve the same result on the adversary's attribution classifier. 
%
%
When they do not, the obfuscated document would evade the  attribution classifier of the obfuscator but not that of the adversary.
%
These differences in the decision boundaries of the two classifiers could be because of the differences in their emphasis on different features; thus, transformations that targeted a certain feature emphasized by one classifier might be rendered useless for the other classifier.
%
%
%
To address this issue, our key insight is that if an obfuscator can evade a meta-classifier, whose decision boundary is based on the decision boundaries of multiple base classifiers, it is more likely to evade those base classifiers.
We hypothesize that a meta-classifier consisting of multiple base classifiers, each emphasizing different features, will better capture the relative importance of various features.
Intuitively, as the internal attribution classifier, this \textit{ensemble} of base classifiers can provide a more nuanced view of the entire feature space and can classify the document in a manner that essentially averages the decision boundaries of the base classifiers.

\begin{figure}
    \includegraphics[width=0.48\textwidth,keepaspectratio]{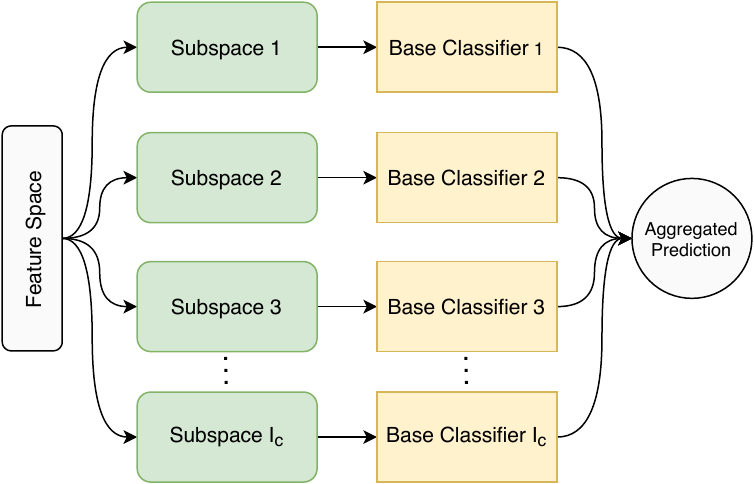}
    \caption{Ensemble architecture for the feature subspacing technique. The original feature space is split into subspaces that are then used to train the base classifiers. The outputs of the base classifiers are then aggregated for the final prediction.}
    \label{fig:ensemble}
\end{figure}

\medskip
\noindent \textbf{Ensemble Approach.}
An ensemble is a learning algorithm that takes a set of classifiers and uses their individual outputs to make the final classification for a given input.
The classifiers in this set are referred to as the base classifiers for the ensemble.
The number of base classifiers affects how the model fits to the training set: too few and the model is likely to underfit and too many will likely result in overfitting.
An efficient number can be determined using cross-validation though it is a time-consuming exercise to train multiple ensembles and validate their results. \cite{kyaw2016determineweaklearners}
The base classifiers can be different classifiers trained on the same training set, or they could be the same and trained on different subsets of the training set (a technique known as \textit{bagging}) or even on subspaces of the feature set \cite{ting2011feature}.

The outputs of the base classifiers are then polled by the ensemble through either a majority vote or by training another classifier (a technique called \textit{stacking}) \cite{ensemble2020dietterich}.
A majority vote gives uniform weight to the output of each base classifier whereas stacking causes the weights to vary, as it can learn to downplay classifiers that are inaccurate more often.
While these base classifiers might not be very accurate on their own, the ensemble together can capitalize on their knowledge and make more accurate predictions.

We construct our ensemble using the feature subspace method and describe it as follows.
A subspace is a subset of the entire universal set of features that are available to the classifier.
We train the base classifiers of the ensemble on different subspaces of the feature set.
The goal of using a subspace of features is to train a base learner that is specialized in that distinct and local set of features.
This is motivated by our findings of feature importance and decision boundaries which we discuss later in Section \ref{sec:discussion}.
The subspaces can be selected randomly \cite{randomsubspace1998kam}, through sampling \cite{stratifiedsampling2013ye}, or through feature selection techniques \cite{greedyfeature2013dyer}.

Figure \ref{fig:ensemble} illustrates the architecture of our proposed ensemble.
The original feature space is divided into multiple subspaces which are then used to train the base classifiers.
The outputs from these base classifiers are then aggregated to produce the final classification of the ensemble for the given input.

%% file: sections/experimental-setup.tex
\section{Experimental Setup}

In this section, we state the assumptions and describe the setup for our experiments.
Specifically, we describe the dataset we use, the attribution classifiers used by the obfuscator and the adversaries, the layout of the experiments, and finally the evaluation metrics used to assess the results.

\input{sections/experimental-setup/data}
\input{sections/experimental-setup/obf-attrib-classifiers}

\input{sections/experimental-setup/adversarial-classifiers}
\input{sections/experimental-setup/design-transfer-exp}
\input{sections/experimental-setup/evaluation-metrics}

%% file: sections/experimental-setup/data.tex
\subsection{Data}

The \textbf{Extended Brennan Greenstadt Corpus} \cite{adversarialStylometrySAfroz} comprises of writing samples submitted by various authors through Amazon's Mechanical Turk (AMT) platform.
The corpus is unique because it was collected expressly for the purpose of adversarial stylometry in text and was vetted against a strict set of guidelines imposed by AMT and the authors themselves to ensure quality.
The guidelines required that the submissions be professionally written, be free of anything other than the writing itself (i.e., citations, URLs, headings, etc.), and contained at least 6500 words.
The imposition of these strict guidelines ensured that the submissions were of high quality, reflected the author's particular writing style, and that there was sufficient data to train an attribution classifier.
Out of the 100 submissions, the authors selected the 45 that most closely followed the guidelines and then split them into nearly 500 word passages, averaging 15 documents per author, to create the final corpus.
In our experiments, we divide the corpus into groups of 5 authors based on document length and report results on these groupings by further splitting them into a 80\% training and 20\% testing set.

%% file: sections/experimental-setup/obf-attrib-classifiers.tex
\subsection{Obfuscator's Attribution Classifiers}

\noindent \textbf{Baseline obfuscator:}
We use \mx{} as the baseline obfuscator for our experiments as its generator only requires black-box access to the attribution classifier. \cite{DBLP:journals/popets/MahmoodASSZ19}
This loose coupling between the two components allows the attribution classifier to be easily swapped for another.
Additionally, we use the Writeprints \cite{abbasi2008writeprints} feature set throughout the experiments to train the internal attribution classifiers for \mx{}.
This feature set incorporates lexical and syntactic features to capture the stylometric properties of the author's writing style.
The lexical features include character-level and word-level features such as total words, average length of words, proportions of different character classes, among others.
The syntactic features include POS tags, use of function words, and various punctuations.

The fitness function for \mx{} takes into account the detection probability of a given attribution classifier $C$ and the semantic similarity between the original and the obfuscated document.
We train the two classifiers that were originally used for \mx{}, a \textit{random forest classifier} and a \textit{support vector machine}, on the Writeprints feature set to serve as baselines for comparing the performance of the ensemble.

\medskip
\noindent \textbf{Writeprints-Static + Ensemble:}
We use the same feature set as our baselines to train the ensemble and construct it by training base classifiers on subspaces of the entire feature set.
We use a linear SVM as the base classifier owing to its stability and demonstrated use as a base classifier for an ensemble in prior work \cite{ting2011feature}.
We use the random subspace method \cite{randomsubspace1998kam} to construct the feature subspaces by randomly choosing distinct features to train each base classifier.
The ensemble then reduces the results from these internal classifiers by polling their individual predictions through a majority vote because it gives a uniform weight to all the base classifiers.
We configure the remaining parameters for the ensemble architecture by conducting small-scale experiments in a variety of settings.
Just as before, we use the training portion of the EBG 5 dataset to select appropriate values for the following hyper-parameters:
\begin{itemize}
    \item number of internal classifiers: $I_c \in \{5, 10, 15\}$
    \item length of subspaces: $L_s \in \{30, 50, 80\}$
\end{itemize}
The results from these experiments show that while a lower value of $L_s$ yields inaccurate individual classifiers, the accuracy of the overall ensemble is much higher.
This ascertains the notion that a robust and highly accurate model can be created from a grouping of weak learners.
In light of this, we conservatively set the value of $L_s = 30$ and $I_c = 10$ as we noticed that higher values yield similar results.
These settings are retained throughout the entirety of our experiments.

%% file: sections/experimental-setup/adversarial-classifiers.tex
\subsection{Adversary's Attribution Classifiers}
\label{sec:adv-clf}

To assess the transferability of the obfuscated samples to other classifiers, we train a series of classifiers and measure the performance of the baselines and the ensemble.
Specifically, we train multiple classifiers that use different types of techniques to measure the \textit{cross-technique transferability} \cite{transferabilityIM2016} of the method used to generate the samples.
These classifiers are also trained on the Writeprints feature set and are as follows: \textit{k-nearest neighbors} (KNN), \textit{naive-bayes} (NB), \textit{multilayer-perceptron} (MLP), \textit{logistic regression} (LR) in addition to the already trained \textit{random forest classifier} (RFC), \textit{support vector machine} (SVM), and the \textit{ensemble} (Ens) itself.
Additionally, we incorporate counter-measures for the findings by Gröndahl et al. \cite{parchoice2020} that using an internal classifier results in highly specific transformations which are not only \textit{non-transferable} to other different classifiers but also to the same classifier if it is retrained.
We accomplish this by training multiple versions of the classifiers that exhibit randomness (RFC, Ensemble, and MLP) during training and then report the average transferability result in their respective columns.

In addition to these Writeprints based classifiers, we also measure the transferability of the samples to JGAAP \cite {juola2009jgaap}, a well-known system for authorship attribution that provides a wide-array of features and classifiers, and another MLP model trained on the Basic-9 feature set\cite{mcdonald2012use}.
This setting allows us to explore the performance of the ensemble against an adversary that uses a different feature set and classifier implementation.
We borrow the configuration recommended by Juola et al \cite{juola2010empirical} for JGAAP.
The final configuration for JGAAP is listed in Table \ref{tbl:jgaap-config}.

\input{tables/tbl-jgaap-config}

%% file: tables/tbl-jgaap-config.tex
\begin{table}[htb]

    \centering
    \begin{tabular}{p{2.45cm}|p{2.45cm}|p{2.45cm}}
         Pre-processing & Features & Classifiers \\
         \midrule
        \RaggedRight Lowercase all
        
        \RaggedRight Strip punctuation
        
        \RaggedRight Normalize whitespaces
        &
         Word unigrams
         
         Word bigrams
         
         Character bigrams
         
         Character trigrams
         
         Function words
         
         Sentence length
        &
        \RaggedRight Weka SMO (SVM)
         
        \RaggedRight Weka Naive Bayes (NB)
         
        \RaggedRight Linear Discriminant Analysis (LDA) \\
    \end{tabular}

\caption{Configuration for JGAAP.}
\label{tbl:jgaap-config}
\end{table}

%% file: sections/experimental-setup/design-transfer-exp.tex
\subsection{Design of Transferability Experiments}

We conduct several experiments with each  
corresponding to a particular internal classifier that is used by \mx{}.
In each of these experiments, all \mx{} parameters are kept consistent and only its internal classifier is replaced.
We configure \mx{} parameters based on the findings from the original pilot experiments performed on the EBG 5 dataset.
The values for the different \mx{} parameters are specified in Table \ref{tbl:mx-config}.
For each experiment, we report the average METEOR score of the obfuscated documents, the transferability rate across our fixed set of adversaries, and the overall attack success rate of that technique.

\input{tables/tbl-mx-config}

In its default setting, \mx{} stops obfuscation once the mutated document has been misclassified by its attribution classifier. 
We alter this behavior to allow \mx{} to continue obfuscation until all $M$ iterations have been performed.
This alteration is partly inspired by the idea that stopping early because of one successful misclassification is detrimental to the overall goal of transferability to a wider set of adversaries.

%% file: tables/tbl-mx-config.tex
\begin{table}
    \centering
    \begin{tabular}{p{0.3\textwidth}|c}
         Setting & Value \\
         \midrule
         Number of word replacements & 5 \\
         Number of iterations & 25 \\
         Number of runs per document & 10 \\
         Weight assigned to attribution confidence in fitness function & 0.75 \\
         Number of top individuals retained in each iteration & 5 \\
         Number of document mutants & 5 \\
    \end{tabular}
    \caption{Configuration of hyper-parameters for \mx{}.}
    \label{tbl:mx-config}
\end{table}

%% file: sections/experimental-setup/evaluation-metrics.tex
\subsection{Evaluation Metrics}

Originally, \mx{} was evaluated through two metrics that measured the safety and soundness of the obfuscation.
While effective for measuring obfuscation across one adversary, they fail to quantify the transferability of the obfuscated samples to multiple adversaries.
To alleviate this, we utilize a third metric called the Attack Success Rate (ASR) to capture this information and slightly modify the safety metric to accommodate it.
The final metrics for evaluation are as follows.

\begin{enumerate}
    \item \textbf{Evasion Effectiveness:}
    An obfuscated document generated from an internal attribution classifier effectively evades an adversary if it is misclassified by that particular adversary.
    We refer to this property of the internal classifier as its \textit{transferability} to an adversary and report it as the percentage of obfuscated documents produced by that classifier that were misclassified by the adversary.
    For an adversary $i$ that misclassifies $a_i$ out of $n$ obfuscated documents generated by the internal classifier, we measure transferability as:
    
    \begin{equation}
        T_i = \frac{a_i}{n}\times100
    \end{equation}
        
    \item \textbf{Attack Success Rate:}
    The attack success rate \cite{regularizedEnsemblesAndTransf2018} measures the overall transferability of the obfuscated documents across the entire set of adversaries. 
    It is an average of all the transferability scores reported for that specific internal classifier.
    Given a fixed set of adversaries $m$, the attack success rate of classifier $A$ can be reported as:
    
    \begin{equation}
        ASR_A = \frac{\sum_{i=1}^{m}T_i}{m}
    \end{equation}

    \item \textbf{Semantic Similarity:}
    An obfuscated document has to maintain semantic similarity to the original document.
    As originally used to evaluate \mx{}, we use the METEOR score \cite{denkowski-lavie-2014-meteor} to assess this similarity.
    The score lies in the range [0, 1] with 1 indicating perfect similarity and 0 indicating the opposite.
    The final score reported is the average METEOR score of the obfuscated documents, where a higher score implies that the final documents were similar to the original ones.
    
\end{enumerate}

%% file: sections/results.tex
\section{Results}

\input{sections/results/evaluation}
\input{sections/results/discussion}

%% file: sections/results/evaluation.tex
\subsection{Evaluation}

\input{tables/tbl-main-results}

The main results of the experiments are presented in Table \ref{tbl:main-results}.
The rows correspond to the internal classifier used by \mx{} and the subsequent columns correspond to the classifier used by the adversary and the feature set they are trained on.
The cell values are the percentage of documents generated using that method that were misclassified by the adversary's classifier, i.e., the transferability of that method.
The final two columns contain the attack success rate (mean of transferability to adversaries) and the mean METEOR score of the technique.
Additionally, we reiterate the counter-measure discussed in Section \ref{sec:adv-clf} and train multiple versions of classifiers that exhibit randomness during training.
As such, the columns for the Ensemble, MLP, and RFC report the average transferability to different versions of the classifiers.

\medskip
\noindent \textbf{Impact of attribution classifier:}
The transferability achieved by SVM ranges from 1.6\% ($SVM \rightarrow RFC$) to 93.7\% ($SVM \rightarrow SVM$), whereas for RFC it ranges from 5.8\% ($RFC \rightarrow LR$) to 29.1\% ($RFC \rightarrow Ensemble$).
In comparison, the ensemble achieves transferability ranging from 15.8\% ($Ensemble \rightarrow LR$) to 71.9\% ($Ensemble \rightarrow Ensemble$).

We notice that the cases where the baselines perform better are where the internal classifier and adversary’s classifier are the same ($SVM \rightarrow SVM, RFC \rightarrow RFC$); cases which are fairly trivial and advantageous to \mx{}.
In fact, the 93.7\% in the case of $SVM \rightarrow SVM$ is unrealistic because the adversary's classifier is exactly the same as \mx{}, whereas $RFC \rightarrow RFC$ and $Ensemble \rightarrow Ensemble$ scenarios have been normalized by training multiple instances of the adversary and reporting their average (see Section \ref{sec:adv-clf}).
Regardless, the ensemble still manages to outperform the other baseline in these trivial cases.

The effects of re-training the ensemble and RFC are also evident in these cases.
The average transferability of $RFC \rightarrow RFC$ is quite low at 28.2\%, corroborating the findings of Gröndahl et al. \cite{parchoice2020}.
However, this doesn't appear to be true for the ensemble as it still reports a fairly high transferability at 71.9\% indicating its robustness.

In the non-trivial cases where the adversary's classifier is different from the internal classifier, the ensemble fares far better than the baselines.
In the case of KNN, the ensemble achieves a transferability of 41.6\% compared to the RFC at 19.4\%.
In the case of NB, it achieves a transferability of 52.9\%, which is 32.3\% higher than the SVM at 20.6\%.
On average, the ensemble achieves 21\% higher transferability than the SVM and 13.8\% higher than the RFC across the set of adversaries where the internal classifier is different.

A comparison of the overall performance (trivial and non-trivial) between the ensemble and the baselines shows that the ensemble outperforms both across the wide range of adversarial settings. 
The overall attack success rate of the ensemble at 38.0\% is the best compared to the baselines, which is 1.7$\times$ higher than the transferability of the RFC at 21.7\% and 2.1$\times$ higher than the transferability of the SVM at 18.3\%.

The ensemble does not perform as well as the other methods when we compare the METEOR scores.
The samples generated by RFC and SVM retain better semantic similarity to the original documents with an average METEOR score of 0.42 and 0.40, respectively.
In contrast, the ensemble reports an average score of 0.36 indicating that the generated samples differed more from their original selves.
We attribute this lower score to the problem of balance between protecting the author's identity and being true to the original content.
We believe that the effort to ensure transferability requires more substantial changes to be made to the document which leads to lower similarity between the source and the obfuscated text, and consequently a lower average METEOR score.

The superior effectiveness of the ensemble as an internal classifier is undeniable when compared to the baselines.
The high attack success rate and a comparable METEOR score make it a reliable alternative to other conventional classifiers for use alongside an obfuscator like \mx{}.

\medskip
\noindent \textbf{Impact of feature set:}
\mx{} may have an inherent advantage when the obfuscator and adversary's classifiers are trained on the same feature set as this likely provides the obfuscator unfair insight into how the adversary operates.
We test this by observing results from experiments where the adversary is trained on a different feature set and classification technique than the internal classifier.

Within the JGAAP setting, the SVM does not perform as well as the other two settings.
It surprisingly performs the worst in the $SVM \rightarrow SVM$ case, only achieving a transferability of 5\%.
We attribute this to the difference between the kernel functions of the two SVMs; as opposed to a linear kernel used in the internal classifier, JGAAP's default setting uses a RBF kernel.
In comparison, the ensemble and the RFC achieve higher degrees of transferability yielding an attack success rate of 34.3\% and 24\% respectively with the ensemble outperforming the RFC by 10.3\%.

We see similar results in the Basic-9 setting, where the ensemble achieves a transferability that is 6\% higher than RFC and almost twice as high the SVM.
This affirms the idea that the ensemble performs just as well against adversaries trained on a different feature set and outperforms other conventional classifiers.

%% file: tables/tbl-main-results.tex
\begin{table*}[t!]
\centering
\begin{adjustbox}{width=1\textwidth}
\begin{tabular}{l|ccccccc|ccc|c|c|c}
        
        \nf \mx  
        & \multicolumn{7}{c|}{Writeprints-Static}
        & \multicolumn{3}{c|}{JGAAP}
        & Basic-9
        & \multirow{2}{*}{ASR}
        & \multirow{2}{*}{METEOR} \\
        Classifier   & RFC*   & SVM   & KNN   & NB & MLP* & LR & Ens* & SVM & LDA & NB & MLP* & & \\ \hline
SVM & \nf 1.6 & \fbox{\nf 93.7} & \nf 18.5 & \nf 20.6 & \nf 10.1 & \nf 1.6 & \nf 7.4 & \nf 5.0 & \nf 15.0 & \nf 10.5 & \nf 16.9 & \nf 18.3 & \nf 0.40 \\
RFC & \fbox{\nf 28.2} & \nf 26.2 & \nf 19.4 & \nf 18.4 & \nf 14.6 & \nf 5.8 & \nf 29.1 & \nf 24.0 & \nf 28.0 & \nf 20.0 & \nf 25.2 & \nf 21.7 & \fbox{\nf 0.42} \\
Ensemble & \nf 18.4 & \nf 61.0 & \fbox{\nf 41.6} & \fbox{\nf 52.9} & \fbox{\nf 21.9} & \fbox{\nf 15.8} & \fbox{\nf 71.9} & \fbox{\nf 32.0} & \fbox{\nf 39.0} & \fbox{\nf 32.0} & \fbox{\nf 31.0} & \fbox{\nf 38.0} & \nf 0.36
 \\

\end{tabular}
\end{adjustbox}
\caption{Transferability results for multiple \mx{} settings. 
Rows are the writeprints-static classifier used for obfuscation.
Columns are the adversary's feature set and classifier.
The asterisk * indicates that the classifier exhibits randomness during training.
Cells underneath a random classifier report the average against multiple instances of that particular classifier.
The attack success rate is summarized in the ASR column and the average METEOR scores are reported in the last column.
}
\label{tbl:main-results}
\end{table*}

%% file: sections/results/discussion.tex
\subsection{Discussion} \label{sec:discussion}

We now study the various decisions we've made concerning the ensemble and try to understand their impact on its inner workings.

\medskip
\noindent \textbf{Impact of ensemble diversity:}
It is widely understood that combining a diverse set of individual classifiers leads to a more robust ensemble \cite{dietterich2000ensembles}.
While the ensemble is diverse in the sense that the base classifiers are trained on distinct subspaces, we instead focus on the predictions of the base classifiers to measure diversity.
Intuitively, the diversity of an ensemble is the difference in the predictions of the individual members that constitute it \cite{Kuncheva2004MeasuresOD}.
To try and understand how the diversity of the ensemble impacts obfuscation, we train multiple ensembles with varying degrees of diversity and compare their performance with each other.

To measure the diversity of an ensemble, we use the non-pairwise metric of entropy $E$ \cite{Kuncheva2004MeasuresOD}.
The entropy value of an ensemble lies in the range $[0, 1]$, a value closer to 0 means that the individual classifiers mostly agree, and a value closer to 1 means that they mostly disagree with one another. 
This measure assumes that a diverse set of classifiers will disagree with one another as opposed to correlated classifiers which will agree more often.

We train multiple ensembles and control for their diversity by selecting appropriate base classifiers.
Specifically, we create 4 bins of entropy values $E \in \{0, 0.25, 0.5, 0.75\}$ and train 10 ensembles for each bin, each ensemble having approximately the same entropy as the bin it is assigned to.
We then conduct 40 experimental runs of \mx{}, and in each run use one of these ensembles as the internal classifier to obfuscate documents and measure the attack success rate of these documents against our set of adversaries.

\begin{figure}
    \centering
    \fbox{\includegraphics[width=0.4\textwidth,keepaspectratio]{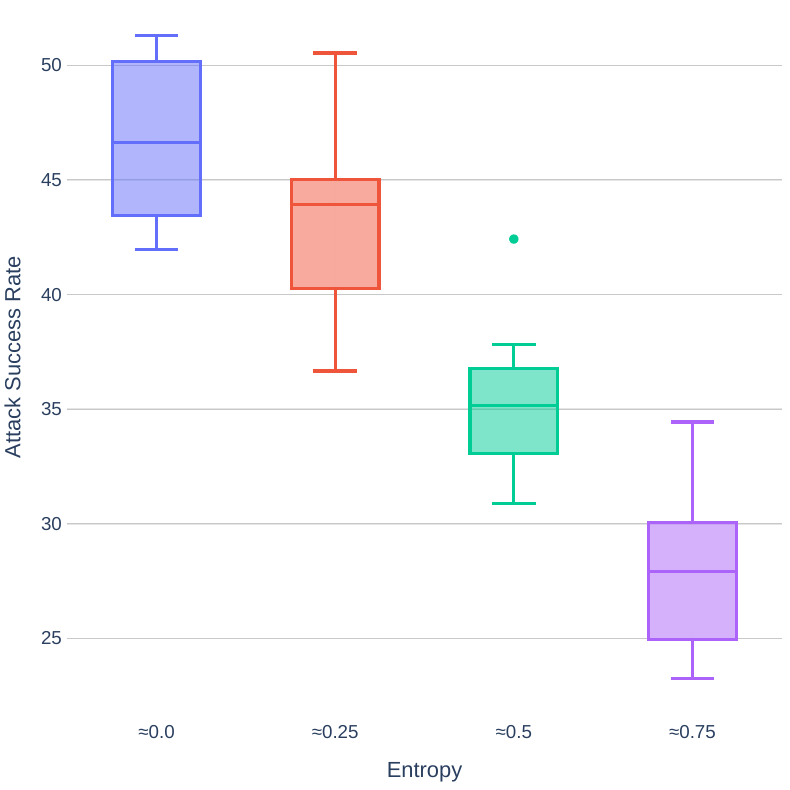}}
    \caption{The attack success rates of the ensembles belonging to each entropy bin. The y-axis represents the attack success rates while x-axis ticks represent the entropy value of the bin. Note the inverse relation between ASR and Entropy: samples sourced from more diverse ensembles fail to transfer well to other adversaries.}
    \label{fig:div-box-plot}
\end{figure}

Figure \ref{fig:div-box-plot} shows the results from these experiments.
The y-axis represents the attack success rates while x-axis ticks represent the entropy value of the bin.
Contrary to our intuition, we notice that ensembles with higher values of entropy (more diverse) had lower transferability while ensembles with lower entropy (less diverse) performed comparatively better.
We explain our interpretation of these results as follows.

Recalling that \mx{} uses the confidence score of the internal classifier to make decisions regarding obfuscation, we investigate the impact of diversity on the accuracy of the ensemble and the confidence of the classifier in its classifications.
To increase the diversity of an ensemble, we need to promote disagreement between the individual classifiers that comprise it.
Consequently, this makes the ensemble overall less confident in its final classification even if it is correct.
Since \mx{} uses this confidence score as an indicator of attribution, a lower score leads to poorer decision making on \mx{}'s part and reduces the quality of obfuscation.
We note that this problem is likely unique to how \mx{} operates and not necessarily an artefact of using ensembles.
In future work, it will be interesting to explore how diversity impacts ensemble based transferability in different settings that are not bound by this restriction.


\begin{figure*}[ht!]
    \centering
    \includegraphics[width=\textwidth,keepaspectratio]{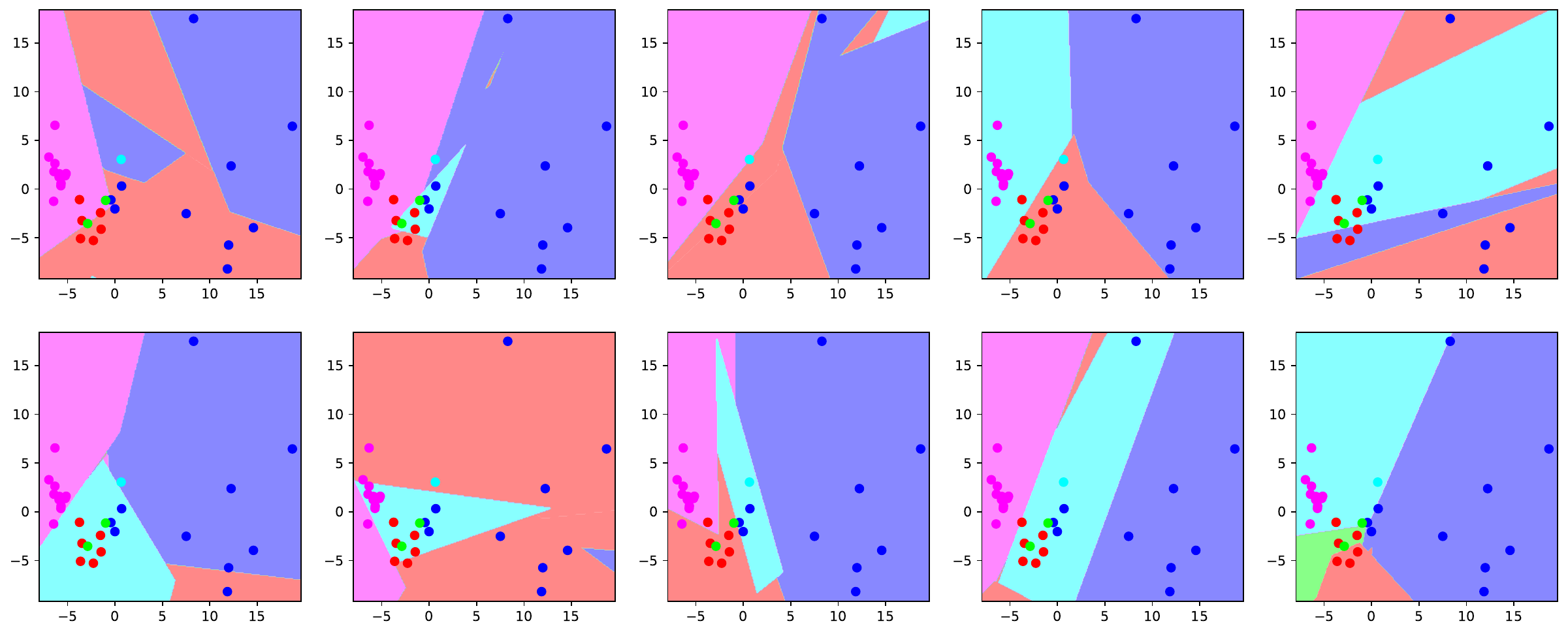}
    \caption{The decision boundaries of the base classifiers inside the ensemble.}
    \label{fig:base-classifier-db}
\end{figure*}

\medskip
\noindent \textbf{Impact of feature subspaces:}
Our main set of experiments consider an ensemble of base classifiers trained on randomly selected subspaces of the feature set.
We now consider more systematic approaches for constructing the subspaces and see how they affect transferability.

At a higher level, the Writeprints feature set incorporates lexical and syntactic features that are qualitatively distinct.
This distinction indicates the presence of a \textit{contextual subspace} within the feature set.
More specifically, there are 9 distinct subspaces and are as follows: \textit{frequency of special characters}, \textit{letters}, \textit{digits}, \textit{parts-of-speech tags} and \textit{punctuation}, \textit{most common letter bigrams and trigrams}, \textit{percentage of certain function words}, and the \textit{ratio of unique words (hapax ratio)} \cite{abbasi2008writeprints}.
We train the base classifiers on this division of subspaces and measure the attack success rate of the resultant ensemble and note that, in contrast to the random subspace method, this yields base classifiers that have different values of $L_s$.

Additionally, we explore feature selection techniques to construct the subspaces.
Using a one-way ANOVA test \cite{elssied2014anova}, we measure the dependency between the features and the author label, and use the highest-ranking features to train the first base classifier.
We repeat this for the next base classifier by considering the remaining set of features and so on for the rest of the classifiers.
This yields base classifiers of varying performance as the initial ones are highly accurate but the accuracy gradually drops as the remaining features are insufficient predictors  .
For this particular experiment, we set $I_c = 8$ and $L_s = 20$ for a consistent distribution of features.

The results of the experiments are as follows: the contextual subspace ensemble yields an ASR of 37.1\% whereas the feature selection subspace ensemble yields an ASR of 34.7\%.
We see that the results for the subspace ensemble are comparable with those of the random subspace ensemble which had an ASR of 38.0\%.

Considering the security aspect of the techniques,
the contextual subspaces approach is relatively more risky. 
Since the features composing these subspaces are easy to identify, 
an adversary can undo the effects of obfuscation through adversarial training: building a classifier to recognize obfuscation by training it on the obfuscated documents generated by the internal classifier  \cite{ehae2014}.
In contrast, an ensemble built using random subspaces offers a good balance: achieving a commensurable degree of transferability and providing a good defence against adversarial training as its random nature is unpredictable for the adversary.


\medskip
\noindent \textbf{Feature importance and decision boundaries:}
While the goal of all classifiers is to map the input document to an author, there are fundamental differences in the way they operate and actually classify the data.
These differences highlight the notion of feature importance: some features are more important for a particular type of classifier than to another.
We now interpret these models to identify the features they consider important and see how it affects transferability.


RFC is a collection of decision trees that also counts the votes of the individual outputs of the trees to make the final classification.
The decision trees consist of several nodes that split the training set into subsets based on the values of certain features.
Arguably, features that are used more often for splitting and split a sizable portion of the training set compared to others bear greater significance for the model.
This is known as the Gini Importance of that feature \cite{breiman2017classification} and it is the number of times a particular feature was used for a split weighted by the number of samples it splits.


SVM classifies the data by learning a hyperplane between the data points that separates the different class boundaries.
In a linear SVM, this hyperplane represents the points at which the distance between the class boundaries is maximum.
Since the coefficients of this hyperplane are associated with the features, their absolute values represent the significance of the corresponding feature relative to the other features.
In a multi-class setting, the SVM has multiple hyperplanes separating each of the classes and each hyperplane has its own set of coefficients.

\input{tables/tbl-feature-importance}

We assess the differences between what features are important for the RFC and SVM.
Table \ref{tbl:feat-importance} lists the top 5 features for the baseline RFC and one of the SVM hyperplanes.
We note that these are different for the two classifiers; moreover, this trend holds even beyond the top 5 features.
In a high-dimensional feature space such as Writeprints, this difference in feature emphasis by the classifier amounts to some features losing their relative importance and thus the obfuscator does not consider their relevance.
This highlights a fundamental flaw in the obfuscator: the obfuscation will always be tuned to the features preferred by its internal classifier and fail to transfer to a different classifier emphasizing different feature.
%


Our approach of using feature subspaces in an ensemble alleviates this flaw to an extent; base classifiers trained on smaller random sets of features emphasize the importance of those features.
The base classifier then \textit{specializes} in its localized subspace of features and, while it may not be accurate, it is representative of a certain aspect of the feature space that might be of significance to an adversary's classifier.

Taking a look at the decision boundaries of the base classifiers in Figure \ref{fig:base-classifier-db} helps explain how they improve transferability.
We use Principal Component Analysis (PCA) \cite{WOLD1987PCA} to reduce the higher-dimensional feature space to two-dimensions for the plotting the boundaries.
The data points are the documents from the test set projected into the PCA dimensions.
The colored regions in the background represent the decision regions of the classifier for that particular label, i.e., points that fall in those regions are classified according to that label.
We stress that this two-dimensional projection is merely an approximation of the actual high-dimensional feature space so some misalignments are expected.
Looking at the decision boundaries of the base classifiers, we see that they vary significantly and that some of the classifiers perform better at classifying a particular author than others.
The decision boundaries also highlight the limited access the classifiers have to the entire feature space, observable by the disjoint between the same decision region.
While the projection takes into account the entire feature space, the decision region is only based on the subspace the particular classifier was concerned with.

\begin{figure}
    \centering
    \includegraphics[width=0.48\textwidth,keepaspectratio]{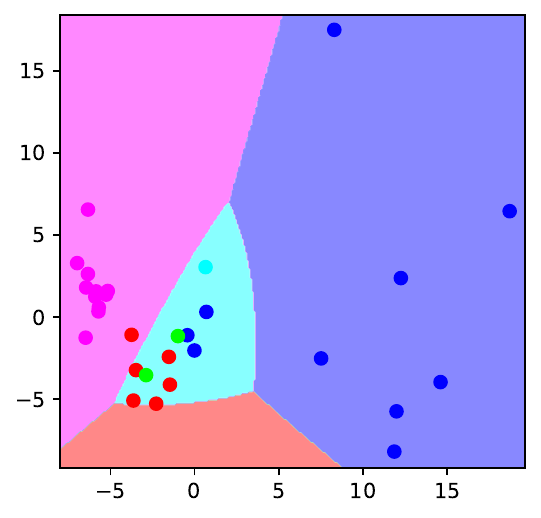}
    \caption{The decision boundary of the ensemble. It provides a middle-ground between the boundaries of the base classifiers and serves as an appropriate attribution classifier.}
    \label{fig:ensemble-db}
\end{figure}

Figure \ref{fig:ensemble-db} shows the decision boundary of the ensemble formed from these base classifiers.
We see that the decision region of the ensemble more closely encapsulates the data points than the base classifiers.
Since the ensemble classifies according to the majority vote of the base classifiers, its decision boundary is approximately the average of all their decision boundaries.
The voting mechanism also ensures that the base classifiers are weighted equally so as to not downplay the role of a certain subspace.
Therefore, the ensemble capitalizes on the individual knowledge of the base classifiers and effectively serves as a \textit{middle-ground} for the obfuscator to compare against.

%% file: tables/tbl-feature-importance.tex
\begin{table}[htb]
    \centering
    
    \begin{adjustbox}{width=0.485\textwidth}
    \begin{tabular}{l|l}
        RFC & SVM \\
        \midrule
        posTagFrequency - Space & functionWordsPercentage - by \\
        frequencyOfLetters - a & functionWordsPercentage - and \\
        functionWordsPercentage - is & functionWordsPercentage - was \\
        frequencyOfDigits - 1 & functionWordsPercentage - in \\
        functionWordsPercentage - had & functionWordsPercentage - that \\
    \end{tabular}
    \end{adjustbox}
    \caption{Top 5 important features for baseline classifiers.}
    \label{tbl:feat-importance}
\end{table}

%% file: sections/related.tex
\section{Related Work}
We survey related research on the transferability of adversarial attacks designed to evade machine learning classifiers.

There is a rich body of literature in the image classification context on transferability of adversarial attacks in both white-box and black-box settings.
Biggio et al. \cite{Biggio2013adversarialexamples} and Szegedy et al. \cite{Szegedy2014adversarialexample} first showed that an adversary can launch attacks by creating minor perturbations in the input that cause machine learning models to misclassify it.
Follow up work has studied the practically of these adversarial attacks in the real world by studying whether they can transfer even when the adversary might not have complete access to the machine learning classifier (e.g., \cite{DBLP:journals/corr/LiuCLS16,Papernot2017blackbox,suciu2018failtransferability,Demontis2019whytransfer}).
For example, Papernot et al. \cite{Papernot2017blackbox} proposed a black-box attack against a variety of machine learning approaches including deep neural networks, logistic
regression, SVM, decision tree, and nearest neighbors, outperforming existing attacks in terms of transferability. 
Suciu et al.
\cite{suciu2018failtransferability} and Demontis et al. \cite{Demontis2019whytransfer} studied if and why adversarial attacks (do not) transfer in real-world settings. 
They showed that the target model's complexity and its alignment with the adversary's source model significantly impact the transferability of adversarial attacks.

Adversarial attacks in the continuous vision/image domain are different than adversarial attacks in the discrete text domain. %
Much of prior work on adversarial attacks is focused on the  vision domain and cannot be easily adapted to the text domain \cite{zhang20adversarialattacknlp}.
Adversarial attacks on text classification models mostly work by simply misspelling certain words \cite{Gao2018adversarialtext,li2010textbugger}.
While these attacks are effective, they are easy to counter by standard pre-processing steps such as fixing misspelled, out-of-vocabulary words. 
Jin et al. (TextFooler) \cite{Jin2020BERTrobust} and Garg et al. (BAE) \cite{garg2020bertadversarialexamples} proposed black-box adversarial attacks on text classification models by replacing certain words using word embeddings and language models, respectively. 
The evaluation showed that these black-box adversarial attacks at best only moderately transfer to unseen models.

Recent adversarial attacks on machine learning based authorship attribution models employing feedback from authorship classifiers are also quite similar.  
Mahmood et al. (\mx{}) proposed a black-box adversarial attack that replaced selected words using word embeddings based on a genetic algorithm \cite{DBLP:journals/popets/MahmoodASSZ19}.
Grondahl et al. \cite{parchoice2020} also proposed a similar black-box adversarial attack (ParChoice) that used paraphrasing to replace selected texts.
While these adversarial attack approaches are effective at authorship obfuscation, they do not transfer well against unseen authorship classifiers. 
Transferable authorship obfuscation in such settings remains an open challenge that we address in our work.

Another relevant line of research has investigated using ensembles to improve transferability of adversarial attacks. 
For example, Liu et al. \cite{DBLP:journals/corr/LiuCLS16} showed that if an adversarial attack succeeds in evading an ensemble of models it will have better transferability because the source ensemble model and the target models are more likely to share decision boundaries.
Most recently, Che et al. \cite{che2020ensembleadversarialattack} studied the effectiveness of different ensemble strategies in improving transferability of adversarial attacks. 
They also conclude that an attack model that evades an ensemble of multiple source models is more likely to transfer to different target models. 

%% file: sections/conclusion.tex
\section{Conclusion}
In the arms race between authorship attribution and  obfuscation, it is crucial that obfuscation can transfer when an adversary deploys a different attributor than the one assumed by the obfuscator. 
In this paper we showed that an ensemble that uses multiple base attribution classifiers, each exploiting random portions of the feature space, is able to achieve better transferability by a factor of 1.7$\times$ and 2.1$\times$.
Moreover, we showed that this success holds even when the adversary's  attributor operates off a different feature space.
We also found that ensemble diversity in terms of disagreement is not crucial for transferability as it only hinders the obfuscator due to a decrease in the ensemble's probability of detection.


%